%% file: main.tex
\documentclass[hidelinks]{article}

\usepackage[final]{corl_2025}

\usepackage{xcolor}
\usepackage{xspace}
\usepackage{amsmath}
\usepackage{amsfonts}
\usepackage[capitalise, nameinlink]{cleveref}
\usepackage{hyperref}
\usepackage[american]{babel}
\usepackage[utf8]{inputenc}
\usepackage[T1]{fontenc}
\usepackage{url}
\usepackage{nicefrac}
\usepackage{microtype}
\usepackage[font=footnotesize]{subfig}
\usepackage[inkscapearea=page]{svg}
\usepackage{mathtools}
\usepackage{graphicx}
\usepackage[labelfont=bf]{caption}
\usepackage{wrapfig}
\usepackage{listings}

\definecolor{codegreen}{rgb}{0,0.6,0}
\definecolor{codegray}{rgb}{0.5,0.5,0.5}
\definecolor{codepurple}{rgb}{0.58,0,0.82}
\definecolor{backcolour}{rgb}{0.95,0.95,0.92}

\lstdefinestyle{mystyle}{
    backgroundcolor=\color{backcolour},   
    commentstyle=\color{codegreen},
    keywordstyle=\color{magenta},
    numberstyle=\tiny\color{codegray},
    stringstyle=\color{codepurple},
    basicstyle=\ttfamily\footnotesize,
    breakatwhitespace=false,         
    breaklines=true,                 
    captionpos=b,                    
    keepspaces=true,                 
    numbers=left,                    
    numbersep=5pt,                  
    showspaces=false,                
    showstringspaces=false,
    showtabs=false,                  
    tabsize=2
}

\author{
\\
\begin{tabular}{cc}
\textbf{Denis Shcherba}\thanks{Equal contribution \ \  Project page: \url{https://mops-tamp.github.io/}} & \textbf{Eckart Cobo-Briesewitz}\footnotemark[1] \\
TU Berlin & TU Berlin \\
\texttt{d.shcherba@campus.tu-berlin.de} & \texttt{cobo-briesewitz@campus.tu-berlin.de} \\
\\
\textbf{Cornelius V.\ Braun} & \textbf{Marc Toussaint} \\
TU Berlin & TU Berlin \\
\end{tabular}
}

\title{Meta-Optimization and Program Search using Language Models for Task and Motion Planning}

\input{macros.tex}

\begin{document}
\maketitle
\begingroup
\endgroup
\begin{abstract}
    Intelligent interaction with the real world requires robotic agents to jointly reason over high-level plans and low-level controls.
    Task and motion planning (TAMP) addresses this by combining symbolic planning and continuous trajectory generation.
    Recently, foundation model approaches to TAMP have presented impressive results, including fast planning times and the execution of natural language instructions.
    Yet, the optimal interface between high-level planning and low-level motion generation remains an open question:
    prior approaches are limited by either too much abstraction (e.g., chaining simplified skill primitives) or a lack thereof (e.g., direct joint angle prediction).
    Our method introduces a novel technique employing a form of meta-optimization to address these issues by: (i) using program search over trajectory optimization problems as an interface between a foundation model and robot control, and (ii) leveraging a zero-order method to optimize numerical parameters in the foundation model output.
    Results on challenging object manipulation and drawing tasks confirm that our proposed method improves over prior TAMP approaches.
\end{abstract}

\keywords{Task and Motion Planning, LLMs as Optimizers, Trajectory Optimization} 

\section{Introduction}
\input{content/intro}


\section{Meta-Optimization and Program Search using Language Models for TAMP}
\input{content/method}

\section{Related Work}
\input{content/related}

\section{Experiments}
\label{sec:Experiments}

\input{content/experiments}

\section{Conclusion}
\input{content/conclusion}

\section{Limitations}
\input{content/limitations}

\acknowledgments{This research was funded by the Amazon Fulfillment Technolgies and Robotics team.}

\bibliography{references}

\newpage
\input{content/appendix}

\end{document}

%% file: macros.tex
  \renewcommand{\a}{\alpha}
  \renewcommand{\b}{\beta}
  \renewcommand{\d}{\delta}

  \renewcommand{\l}{\lambda}
  \renewcommand{\L}{\Lambda}
    \newcommand{\m}{\mu}

  \renewcommand{\k}{\kappa}
  \renewcommand{\o}{\omega}
  \renewcommand{\O}{\Omega}

  \renewcommand{\r}{\varrho}
    \newcommand{\s}{\sigma}
  
  \renewcommand{\t}{\theta}

    \newcommand{\CC}{{\cal C}}

    \newcommand{\JJ}{{\cal J}}

  \renewcommand{\SS}{{\cal S}}

  \newcommand{\RRR}{{\mathbb{R}}}

  \newcommand{\vx}{\mathbf{x}}

  \renewcommand{\[}{\Big[}

  \renewcommand{\|}{\,|\,}
  
  \renewcommand{\=}{\!=\!}

  \renewcommand{\vec}{\boldsymbol}

  \providecommand{\href}[2]{{\color{blue}USE PDFLATEX!}}
  \providecommand{\url}[2]{\href{#1}{{\color{blue}#2}}}
  \newcommand{\anchor}[2]{\begin{picture}(0,0)\put(#1){#2}\end{picture}}

  \newcommand{\fullhide}[1]{\message{^^JHIDE--Warning (hidden)!^^J}}
  \newcommand{\mytitle}[1]{\title{#1}\newcommand{\thetitle}{#1}}
  \newcommand{\header}{\begin{document}\mytitle\cleardefs}
  \newcommand{\contents}{{\tableofcontents}\renewcommand{\contents}{}}
  \newcommand{\footer}{\small\bibliography{marc,bibs}\end{document}}
  \newcommand{\widepaper}{\usepackage{geometry}\geometry{a4paper,hdivide={25mm,*,25mm},vdivide={25mm,*,25mm}}}
  \newcommand{\moviex}[2]{\movie[externalviewer]{#1}{#2}} 
  \newcommand{\rbox}[1]{\fboxrule2mm\fcolorbox[rgb]{1,.85,.85}{1,.85,.85}{#1}}
  \newcommand{\mpage}[2]{{\begin{minipage}{#1\columnwidth}#2\end{minipage}}}
  \newcommand{\redbox}[2]{\fboxrule1mm\fcolorbox[rgb]{1,.7,.7}{1,.7,.7}{\begin{minipage}{#1\columnwidth}\center#2\end{minipage}}}
  \newcommand{\onecol}[2]{
    \begin{minipage}[c]{#1\columnwidth}#2\end{minipage}}
  \newcommand{\twocol}[5][0]{
    \begin{minipage}[c]{#2\columnwidth}#4\end{minipage}\hspace*{#1\columnwidth}%
    \begin{minipage}[c]{#3\columnwidth}#5\end{minipage}}
  \newcommand{\threecol}[7][0]{%
    \begin{minipage}[c]{#2\columnwidth}#5\end{minipage}\hspace*{#1\columnwidth}%
    \begin{minipage}[c]{#3\columnwidth}#6\end{minipage}\hspace*{#1\columnwidth}%
    \begin{minipage}[c]{#4\columnwidth}#7\end{minipage}}
  \newcommand{\threecoltext}[7][c]{
    \begin{minipage}[#1]{#2\textwidth}#5\end{minipage}%
    \begin{minipage}[#1]{#3\textwidth}#6\end{minipage}%
    \begin{minipage}[#1]{#4\textwidth}#7\end{minipage}}
  \newcommand{\threecoltop}[7][0]{%
   \begin{minipage}[t]{#2\columnwidth}#5\end{minipage}\hspace*{#1\columnwidth}%
   \begin{minipage}[t]{#3\columnwidth}#6\end{minipage}\hspace*{#1\columnwidth}%
   \begin{minipage}[t]{#4\columnwidth}#7\end{minipage}}
  \newcommand{\fourcol}[9][0]{%
   \begin{minipage}[c]{#2\columnwidth}#6\end{minipage}\hspace*{#1\columnwidth}%
   \begin{minipage}[c]{#3\columnwidth}#7\end{minipage}\hspace*{#1\columnwidth}%
   \begin{minipage}[c]{#4\columnwidth}#8\end{minipage}\hspace*{#1\columnwidth}%
   \begin{minipage}[c]{#5\columnwidth}#9\end{minipage}}
  \newcommand{\helvetica}[4]{\setlength{\unitlength}{1pt}\fontsize{#1}{#1}\linespread{#2}\usefont{OT1}{phv}{#3}{#4}}
  \newcommand{\helve}[1]{\helvetica{#1}{1.5}{m}{n}}
  \newcommand{\german}{\usepackage[german]{babel}\usepackage[utf8]{inputenc}}
\newcommand{\sans}{\renewcommand{\familydefault}{\sfdefault}}

\newcommand{\norm}[1]{|\!|#1|\!|}
\newcommand{\expr}[1]{[\hspace{-.2ex}[#1]\hspace{-.2ex}]}

\newcommand{\Jwi}{J^\sharp_W}
\newcommand{\THi}{T^\sharp_H}
\newcommand{\Jci}{J^\natural_C}
\newcommand{\hJi}{{\bar J}^\sharp}
\renewcommand{\|}{\,|\,}
\renewcommand{\=}{\!=\!}
\newcommand{\myminus}{{\hspace*{-.0pt}\text{\rm -}\hspace*{-.5pt}}}
\newcommand{\myplus}{{\hspace*{-.0pt}\text{\rm +}\hspace*{-.5pt}}}
\newcommand{\1}{{\myminus1}}
\newcommand{\2}{{\myminus2}}
\newcommand{\3}{{\myminus3}}
\newcommand{\mT}{{\text{\rm -}\hspace*{-1pt}\top}}
\newcommand{\po}{{\myplus1}}
\newcommand{\pt}{{\myplus2}}
\newcommand{\T}{{\!\top\!}}
\newcommand{\xT}{{\underline x}}
\newcommand{\uT}{{\underline u}}
\newcommand{\zT}{{\underline z}}
\newcommand{\Sum}{\textstyle\sum}
\newcommand{\Int}{\textstyle\int}
\newcommand{\Prod}{\textstyle\prod}

\newenvironment{centy}{
\vspace{15mm}
\large
\hspace*{5mm}
\begin{minipage}{8cm}\it\color{blue}
}{
\end{minipage}
}

\newcommand{\old}{{\text{old}}}
\newcommand{\new}{{\text{new}}}
\newcommand{\MAP}{{\text{MAP}}}
\newcommand{\ML}{{\text{ML}}}

\newcommand{\redArrow}{\quad\anchor{0,-1}{\includegraphics[scale=.5]{figs/redArrow}}}
\newcommand{\pub}[1]{{\color{green}\helvetica{8}{1.3}{m}{n}#1\\}}
\newcommand{\KL}[2]{\opKL\big(#1\,\big|\!\big|\,#2\big)} 

\renewcommand{\show}[2][.7]{\centerline{\includegraphics[width=#1\columnwidth]{#2}}}
\newcommand{\showh}[2][.8]{\includegraphics[width=#1\columnwidth]{#2}}
\newcommand{\shows}[2][.8]{\centerline{\includegraphics[scale=#1]{#2}}}
\newcommand{\showhs}[2][.8]{\includegraphics[scale=#1]{#2}}
\newcommand{\mov}[2]{\movie[externalviewer]{{\color{blue}\small #1}}{movies/#2}}
\newcommand{\movex}[2]{\movie[externalviewer]{#1}{#2}} 
\newcommand{\movh}[3][autostart,loop]{
\movie[#1]{\showh[#2]{#3.jpg}}{#3.mp4}%
\movie[externalviewer]{$\circ$}{#3.mp4}
}
\newcommand{\movc}[3][autostart,loop]{\centerline{\movh[#1]{#2}{#3}}}
\newcommand{\mymovie}[3][]{\centerline{\movie[autostart,loop,#1]{\includegraphics[#1]{#2}}{#3}}}

\newcommand{\cen}[1]{\centerline{#1}}

\newcommand{\citing}[1]{
{\color{citcol}\ttiny#1\par}
}

\newcommand{\cit}[3]{
\par\smallskip
{\color{citcol}\ttiny #1: \emph{#2}. #3 \par}
}

\newcommand{\citurl}[4]{
\par\smallskip
{\color{citcol}\ttiny #1: \protect{\href{#4}{\color{blue}{#2.}}} #3 \par}
}

\newcommand{\cito}[3]{
\par\smallskip
{\color{bluecol}\ttiny #1: \emph{#2}. #3 \par}
}

\newcommand{\redoMacrosInProof}{
  \renewcommand{\d}{\delta}
  \renewcommand{\=}{\!=\!}
}

\newcommand{\pdflatex}{
  \definecolor{bluecol}{rgb}{0,0,.5}
  \usepackage[
    colorlinks,
    urlcolor=bluecol,
    citecolor=black,
    linkcolor=bluecol,
    pdfborder={0 0 0},
    pdfpagemode=UseOutlines, 
    pdfauthor={Marc Toussaint}
  ]{hyperref}
  \DeclareGraphicsExtensions{.pdf,.png,.jpg,.eps}
  \renewcommand{\r}{\varrho}
  \renewcommand{\l}{\lambda}
  \renewcommand{\L}{\Lambda}
  \renewcommand{\s}{\sigma}
  \renewcommand{\b}{\beta}
  \renewcommand{\d}{\delta}
  \renewcommand{\k}{\kappa}
  \renewcommand{\t}{\theta}
  \renewcommand{\O}{\Omega}
  \renewcommand{\o}{\omega}
  \renewcommand{\SS}{{\cal S}}
  \renewcommand{\=}{\!=\!}
}

\newcommand{\mypause}{\pause}
\newcommand{\defi}[1]{\textbf{#1}}
\newcommand{\red}[1]{\emph{\color{red}#1}}
\newcommand{\pos}{{\textsf{pos}}}
\newcommand{\eff}{{\textsf{eff}}}
\newcommand{\rot}{{\textsf{rot}}}
\newcommand{\veC}{{\textsf{vec}}}
\newcommand{\quat}{{\textsf{quat}}}
\newcommand{\col}{{\textsf{col}}}
\newcommand{\de}[2]{\frac{\partial #1}{\partial #2}}
\newcommand{\target}{{\text{target}}}
\newcommand{\near}{{\text{near}}}
\newcommand{\qfree}{Q_{\text{free}}}
\renewcommand{\vec}{\boldsymbol}
\newcommand{\lft}{\text{left}}
\newcommand{\rgh}{\text{right}}
\newcommand{\prev}{{\text{prev}}}
\newcommand{\TR}[2]{T_{{#1}\to{#2}}}
\newcommand{\RO}[2]{R_{{#1}\to{#2}}}
\newcommand{\liter}{\helvetica{8}{1.1}{m}{n}\parskip 1ex}
\newcommand{\Fc}{\color{green}F}
\newcommand{\muc}{\color{blue}\mu}
\newcommand{\Astar}{A$^*$}

\providecommand{\defn}[1]{\textbf{#1}}

\newcommand{\adec}{\r_\a^-}
\newcommand{\ainc}{\r_\a^+}
\newcommand{\ldec}{\r_\l^-}
\newcommand{\linc}{\r_\l^+}
\newcommand{\minc}{\r_\m^+}
\newcommand{\mdec}{\r_\m^-}
\newcommand{\step}{{\delta}}
\newcommand{\lsstop}{\r_{\text{ls}}}
\newcommand{\stepmax}{\step_{\text{max}}}
\newcommand{\Min}{\text{min}}
\newcommand{\Max}{\text{max}}
\DeclareMathOperator*{\argmin}{arg\,min}
\DeclareMathOperator*{\argmax}{arg\,max}
\DeclareMathOperator*{\EEE}{\mathbb{E}}

\definecolor{amaranth}{rgb}{0.9, 0.17, 0.31}
\definecolor{BLUE}{rgb}{0,0,1}
\definecolor{algblue}{HTML}{125DA6} 
\newcommand{\rred}[1]{\textcolor{amaranth}{#1}}
\newcommand{\bblue}[1]{\textcolor{algblue}{#1}}

\newcommand{\wrt}{w.r.t.\@\xspace}
\newcommand{\cma}{\Delta \vx_{\textsc{cma}}}

\newcommand{\todo}[1]{{\color{red}\textbf{todo: #1}}}
\newcommand{\marc}[1]{{\color{blue}\textbf{marc: #1}}}
\newcommand{\added}[1]{{\color{olive}\textbf{#1}}}

%% file: content/intro.tex
Intelligent interaction with the world requires flexible action plans that are robust and satisfy real-world constraints.
To achieve a long-horizon goal, agents are required not only to reason over symbolic decisions, but also to make geometry-based decisions to seamlessly integrate high-level reasoning and low-level control.
The resulting decision-making problems are challenging to solve due to the combinatorial complexity of the search space with increasing plan lengths.

Commonly, the challenges of these decision-making problems are approached through the lens of integrated task and motion planning (TAMP) \cite{garrett2021integrated, toussaint2015logic, kaelbling2011hierarchical, lozano2014constraint}.
The core idea of TAMP is to make long-horizon sequential manipulation planning tractable by introducing a symbolic domain that links discrete high-level decisions with continuous motions \cite{danny2024}.
Over the past decades, this framework has been shown capable of solving a wide variety of tasks -- from table-top manipulation \cite{xue2024d, srivastava2014combined, braun2022rhh}, to puzzle solving \cite{levit2024solving}, to architectural construction \cite{hartmann2020robust}.

While highly general, classical TAMP approaches trade expressivity for tractability: they fix the mapping between symbolic skills and the underlying trajectory optimization, a function that must be designed by an expert engineer.
This imposes two key limitations: First, solution quality depends on the designer's insight and experience. 
Second, the flexibility of these methods is severely limited because the reasoning over NLPs occurs only at the symbolical level, while the specific timings, scalings, and other parameters of the trajectory constraints are abstracted away during planning. 
This rigid separation often leads to plans that are only locally optimal or not feasible at all \cite{garrett2021integrated, curtis2024trust}.

\begin{figure}[t]
    \centering
    \begin{minipage}[t]{0.77\textwidth}
        \vspace{0pt}  
        \includegraphics[page=3, width=\textwidth, trim={2cm 2cm 1cm 1cm},clip]{imgs/mops_plots.pdf}
    \end{minipage}\hfill
    \begin{minipage}[t]{0.23\textwidth}
        \vspace{0pt}  
        \centering
        \includegraphics[width=\textwidth, trim={0cm 0cm 0cm 0.1cm},clip]{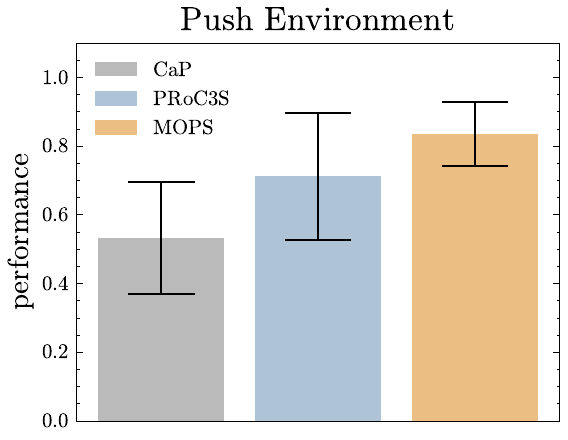}\\[1ex]
        \includegraphics[width=\textwidth, trim={0cm 0cm 0cm 0.1cm},clip]{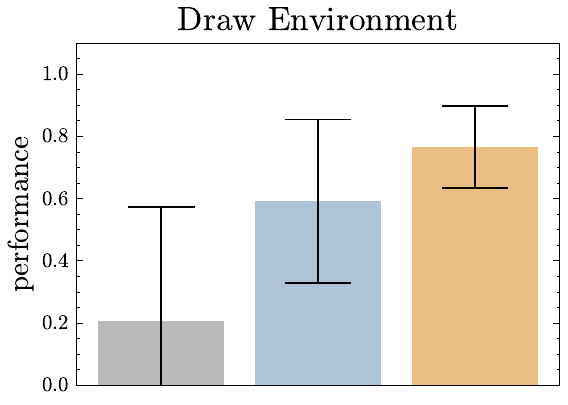}
    \end{minipage}
    \caption{Overview diagram of our method MOPS and its empirical performance. \textbf{Left:} MOPS solves the problem by iterating a meta-optimization loop: an LLM optimizes the selection of constraints, a blackbox optimizer (BBO) improves the continuous parameters of these functions and a gradient-based method solves the induced non-linear program (NLP) for a trajectory, which is then simulated to compute the full cost. Based on the back-propagated costs, the constraints of the NLP are adapted until convergence. \textbf{Right:} Average normalized performance across two domains, with three tasks each. MOPS outperforms prior methods that search over action sequences (PRoC3S) or simple code snippets (CaP). Results for each task are averaged over 5 independent runs ($\pm1.96$ standard deviation). }
    \label{fig:visualAbstract}
\end{figure}

%
Recently, foundation model (FM)-based methods have been presented as an alternative to classical TAMP methods, since they do not require hand-crafted symbolic predicates to solve tasks. 
Instead, they can select control policies from a pre-trained repertoire \cite{driess23a}, write code that controls the robot \cite{liang2023code}, or directly predict joint states \cite{wang2024prompt} given a high-level goal specification.

While FMs have now been applied successfully to a wide range of tasks, most prior approaches are built on simplifying and restrictive assumptions. 
A common limiting factor is their reliance on atomic skills, originating from the lack of fine‑grained spatial reasoning in current foundation models \cite{liang2023code, singh2023progprompt, brohan2023can, huang2023grounded}.
This allows them to solve simple pick-and-place tasks, but fails for tasks that require precise placement locations or low tolerances for errors.
\citet{curtis2024trust} address this shortcoming by formulating language-instructed TAMP as a constraint satisfaction problem. 
To solve this problem, the authors propose using LLMs to reason over parameterized skills, for which the parameters can be sampled so that the specified constraints hold during manipulation.
However, their approach requires human users to specify relevant task constraints and uses uniform sampling to obtain skill parameters.
The result is sample‑inefficient rejection sampling \cite{casella2004generalized}, and an approach that heavily hinges on the capabilities of the user to specify all relevant constraints upfront, which can be challenging for many real-world tasks.

In this paper, we introduce \textit{Meta-Optimization and Program Search using Language Models for TAMP (MOPS)}, a novel approach that overcomes these limitations by treating the TAMP problem as a \textit{meta-optimization} problem in which both the motion constraints and the resulting trajectories are explicit optimization targets.
Concretely, we solve this problem by nesting three optimization levels.
In the first step, we use an FM as an optimizer \cite{song2024position} to propose parameterized constraint functions $c(x, \alpha_c)$ that define a nonlinear mathematical program (NLP) and sensible initialization heuristics for each numerical constraint parameter in a plan.
Second, we perform black-box optimization to optimize the numerical constraint parameters $\alpha_c$ for overall task success and reward. 
Third, we perform gradient-based trajectory optimization. This step solves the fully defined NLP to produce smooth, and collision-free trajectories that satisfy all task constraints $c(x, \alpha_c)$.

Our contributions are summarized as follows:
\begin{itemize}
    \item We propose a novel perspective on TAMP by formulating language conditioned TAMP as a search over constraint sequences instead of actions sequences (\cref{secProblem}).
    \item We introduce a novel multi-level optimization method for sequential robotic manipulation that combines foundation models with parameterized NLPs and gradient-based trajectory optimization, which enables efficient complex robot manipulation (\cref{secMops}).
    \item We validate our method, MOPS, on a range of problems and demonstrate that it improves over prior TAMP approaches (\cref{secExp}).
\end{itemize}

%% file: content/method.tex
This section introduces our method, MOPS (Meta-Optimization and Program Search), a hierarchical framework that casts language conditioned TAMP as a meta-optimization problem.
At the top level, a foundation model (FM) performs a semantic search over discrete constraints; at the middle level, a black‐box optimizer refines continuous constraint parameters; and at the lowest level, a gradient‐based solver produces a smooth trajectory.

\subsection{Problem Formulation}\label{secProblem}
We start by defining the language-conditioned TAMP setting, which follows the general TAMP problem structure \cite{garrett2021integrated}, but assumes that the task goal is specified in natural language instead of a planning language such as PDDL.

\begin{figure}[t]
  \centering
  \subfloat[Environment State]{%
    \raisebox{10pt}[\height][\depth]{
      \includegraphics[page=1,width=0.2\textwidth,trim={0cm 6cm 42cm 2cm},clip]{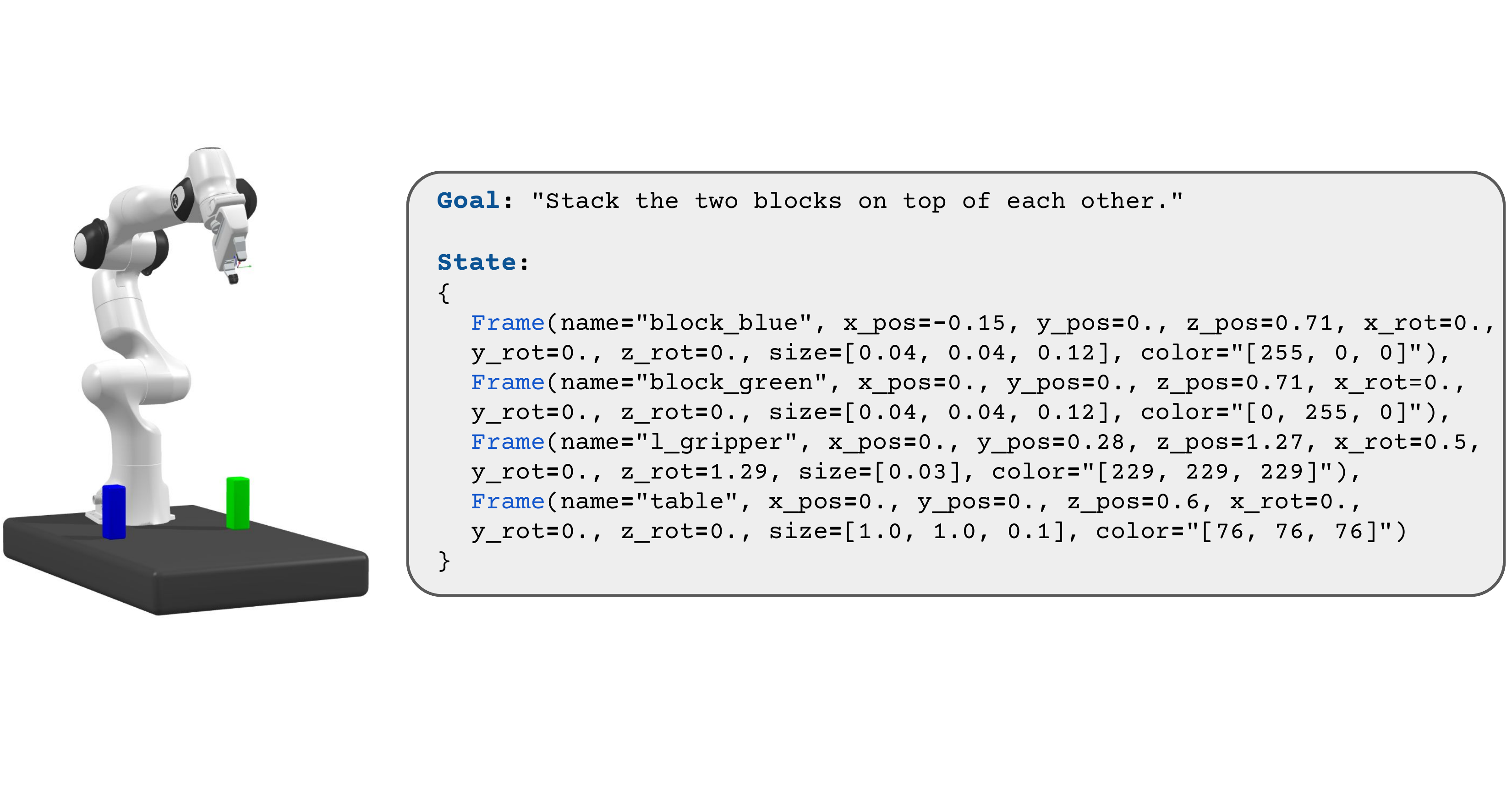}%
    }%
  }\hfill
  \subfloat[User LLM prompt]{%
    \raisebox{0pt}[\height][\depth]{
      \includegraphics[page=1,width=0.7\textwidth,trim={15cm 6cm 0cm 6cm},clip]{imgs/mops_plots.pdf}%
    }%
  }
  \caption{Illustration of the state definition and user goal description that we prompt the LLM with. (a) Visualizes the environment state, and (b) the corresponding prompt of the LLM, specifying the initial state $s_0$ as Python dictionary and the planning goal $G$ in natural language.}
  \label{fig:state_vis}
\end{figure}

A language‐conditioned TAMP problem is specified by the tuple $(\SS, \CC, \,s_0,\,G,\,\JJ,\,g,\,h)$.
Here, $\SS$ denotes the fully observable state space, which includes robot end effector, and object poses, as well as geometry dimensions and colors (see Fig.\ \ref{fig:state_vis} for an example).
Further, $\CC = \RRR^{n} \times \text{SE}(3)$ denotes the configuration space of the scene with an $n$-dof robot, and $s_0$ denotes the initial state of the task which has natural language goal $G$.

We formulate the language-conditioned TAMP problem following prior work that translates TAMP into constrained optimization problems \cite{toussaint2015logic}.
Given a natural-language goal $G$, the objective is to optimize two types of decision variables: \textit{(i)} the continuous trajectory parameters $x \in \RRR^{T\times n}$, and \textit{(ii)}, the mixed-integer constraint parameters $\alpha \in [0, 1]^k \times \RRR^j$.
Specifically, $\alpha=[\alpha_i,\alpha_c]$ consists of binary selectors $\alpha_i\in\{0,1\}^{k}$ indicating \emph{which} constraints from a predefined library shall be enforced, and continuous parameters $\alpha_c\in\RRR^{j}$ specifying their numerical values (e.g., grasp or placement poses).
This approach stands in stark contrast to classical TAMP solvers.
While those search over a sequence of \textit{actions} (e.g., relating to a PDDL), our approach uses LLMs to search over a set of \textit{constraints} that specify a motion NLP.
Following the TAMP and trajectory optimization literature, we denote inequality constraint maps by $g$ and equality constraints by $h$ \cite{toussaint2015logic}. 
For fixed $\alpha$, the constraint functions are $g(\cdot,\alpha):\RRR^{T\times n}\!\to\RRR^{m_\alpha}$ and $h(\cdot,\alpha):\RRR^{T\times n}\!\to\RRR^{p_\alpha}$, where $m_\alpha,p_\alpha$ depend on $\alpha$. 
Both $g$ and $h$ can be nonlinear and nonconvex constraints, and we only assume that these maps are (piecewise) continuously differentiable in $x$.

The full optimization objective is to find a set of plan parameters $x, \alpha$, such that the overall task cost $\JJ$ is minimized and that all task constraints are satisfied.
Formally, the goal is to optimize
\begin{equation}
  \min_{x,\alpha} \JJ(x,\alpha)
  = \int_0^T \bigl[f(x(t),\alpha) + \Psi(x(t),\alpha)\bigr]\,\mathrm{d}t \label{eq:tamp}
\end{equation}
\begin{equation}
  \begin{aligned}
    \text{s.t.}\quad
    g\bigl(x(t),\alpha\bigr) &\le 0,\quad \forall t\in[0,T],\\
    \phantom{\text{s.t.}\quad}
    h\bigl(x(t),\alpha\bigr) &= 0,\quad \forall t\in[0,T].
  \end{aligned}
\end{equation}
Following the trajectory optimization literature, we assume that $\JJ$ is a linear combination of continuous, and differentiable trajectory costs $f(x, \alpha)$, and an \textit{extrinsic} cost $\Psi(x, \alpha)$ which we treat as a black box.
In practice, we use $f(x(t), \alpha) = \ddot x(t)^T\ddot x(t)$, i.e., we minimize squared accelerations. 
The extrinsic costs $\Psi$ quantify the degree of task success of a trajectory solution, and depend on the domain (we release our code including all cost functions in the supplementary materials).
By decoupling the optimization of $f$ and $\Psi$, we can exploit specialized solvers for each term.

\subsection{The Challenges of Language-Conditioned TAMP}
Leveraging FMs for TAMP introduces two key difficulties:
First, it is well known that foundation models struggle at geometrical and numerical reasoning \cite{liang2023code, curtis2024trust, driess23a}.
As a result, it is commonly infeasible to synthesize low-level control signals directly from the foundation model.
To address this issue, one needs to introduce abstraction layers to the problem.
In our case, we take inspiration from recent work by \citet{curtis2024trust} in separating discrete constraint sampling and continuous parameter sampling.
Second, the optimization problem in Eq.\ \eqref{eq:tamp} is difficult to solve due to its mixed integer structure, black-box extrinsic cost $\Psi$ and high dimensionality.
In particular, since many constraints can be selected, each of which depends on multiple parameters, optimizing all variables concurrently is infeasible.
For instance, consider the pushing task depicted in Fig.\ \ref{fig:visualAbstract}.
In this task, the robot must not only use the correct set of trajectory constraints to enforce the desired motion, but it also must specify multiple specific poses overtime to fully specify a successful push.
We therefore propose to solve the problem by performing a multi-level meta-optimization, which solves each part of the problem leveraging an appropriate method.

\subsection{MOPS: Breaking Down the Problem into Three Levels}\label{secMops}
MOPS (Meta-Optimization and Program Search) frames the language-conditioned TAMP problem of Eq.\ \ref{eq:tamp} as a meta-optimization over a Language Model Program \cite[LMP]{liang2023code}, i.e., a parameterized non-linear program (NLP).
We denote such a program by $\text{NLP}(x, \alpha)$ in the following.
Rather than solving discrete constraint selection, continuous parameter optimization, and trajectory synthesis in a single step, MOPS interleaves them in an iterative loop: 
a foundation model proposes $\alpha_i$ and an initial guess $\alpha_c^{init}$; a black-box optimizer refines $\alpha_c$ via simulation-based evaluations of the extrinsic cost $\Psi$; and a gradient-based solver computes the smooth trajectory $x$ under the instantiated constraints. 
By repeating these complementary stages, MOPS jointly refines both the structure and the parameters of the NLP, driving down the overall cost $\JJ$. 
The full approach is illustrated in Fig.\ \ref{fig:visualAbstract}.
We now describe each component in detail.

\paragraph{Level 1: Language Model Program Search.}
The goal of this stage is to optimize the set of constraints $\alpha_i$ that shall be enforced in the NLP.
Thus, the objective is to find the discrete selector $\alpha_i$ that minimizes the extrinsic cost, i.e.,
\begin{equation}
    \min_{\alpha_i} \text{Cost}(\alpha_i) = \Psi(x, \alpha_i, \alpha_c),
\end{equation}
To generate the optimal NLP \wrt $\alpha_i$, we prompt a foundation model with a textual description of the state space $\SS$, the initial state $s_0$, the available constraint functions, and two in-context examples for a different task (we list our prompts in \cref{secPrompts}).
The model is then asked to return: First, a parameterized NLP generation function, which implements constraints selected by the discrete variables $\alpha_i \subset \alpha$.
This NLP can be solved for a trajectory $x$ once it is fully parameterized by its continuous parameters $\alpha_c$. 
Second, the FM is tasked to return an initial guess $\alpha_c^{init}$ for the continuous parameters $\alpha_c$, which we further optimize at level two.
This prompt structure roughly follows that of \citet{curtis2024trust}, but instead of generating the bounds of a uniform sampling domain, we query the FM for an initial guess for the optimizer.

\paragraph{Level 2: Constraint Parameter Optimization.}
At this stage, we optimize the continuous constraint parameters of the NLP in Eq. \ref{eq:tamp}, given a set of constraint selection $\alpha_i$ from level 1.
To perform this optimization, we iteratively evaluate each set of NLP parameters $\alpha = [\alpha_i, \alpha_c]$ by solving the NLP for the motion $x$.
Subsequently, we execute the solution in simulation to evaluate $\text{Cost}(\alpha_c) = \Psi(x, \bar{\alpha}_i, \alpha_c)$, where $\bar{\alpha}_i$ indicates that $\alpha_i$ is fixed at this level.
Since the non-differentiable task-cost $\Psi$ is commonly fast to evaluate, we can use a generic black-box optimizer  \cite{hansen2003reducing} to optimize the continuous trajectory optimization parameters $\alpha_c$ by iteratively solving the corresponding NLPs at the lowest level.

\paragraph{Level 3: Gradient-Based Trajectory Optimization.}
The goal of this step is to optimize the robot trajectory $x$ by solving a fully parameterized NLP.
Given $\alpha=[\alpha_i, \alpha_c]$, we solve
\begin{equation}
    \min_x \int_0^T f(x(t),\alpha)\,\mathrm{d}t \qquad \text{s.t.} \qquad g(x(t), \alpha) \leq 0, h(x(t), \alpha) = 0~ \forall t
\end{equation}
using a second-order Augmented Lagrangian method to produce a smooth joint-state trajectory $x$.
We then roll out $x$ inside a physical simulator to obtain the full task cost $\JJ(x, \alpha)$, which is then used to further refine the higher-level decision variables $\alpha$.

\paragraph{Closing the Loop.}
We repeat these three levels for a specified number of steps, or until the optimization has converged.
Closing the loop enables us to optimize the discrete constraint set selection $\alpha_i$ using the FM.
To achieve this, we provide the foundation model with feedback about the outcome of the lower level stages.
In practice, we return information about \textit{(i)} the lowest trajectory cost that was found, \textit{(ii)} the final state after running the trajectory, and \textit{(iii)} a target cost that should be achieved for convergence.
Closing this loop enables to leverage the FM as a black-box optimizer, as it continuously improves constraint proposals based on prior cost values.

%% file: content/related.tex

\paragraph{Task and Motion Planning.}
    The field of Task and Motion Planning (TAMP) \cite{garrett2021integrated} has generated various powerful methods for completing complex manipulation tasks in zero-shot manner.
    What unifies all TAMP approaches is that they solve the task by combining symbolic and discrete reasoning and continuous motion planning \cite{danny2024, toussaint2015logic, lozano2014constraint}.
    While this approach is highly general, it requires the definition of a fixed set of predicates for reasoning, which can be very challenging in practice \cite{silver2021learning}.
    A further challenge in TAMP is the combinatorial complexity of the search space during planning.
    Therefore, recent work resorted to learning-based approaches \cite{driess2020deep, fang2024dimsam, silver2021learning, silver2023predicate}.
    In particular, multiple works replaced traditional search-based planners by LLMs \cite{silver2024generalized, ahn2022can, curtis2024trust}.
    Our work follows this approach, as we also use a foundation model to optimize the task plan.

\paragraph{Foundation Models in Robotics.}
LLMs are increasingly applied to robot control and planning \cite{wang2024large, firoozi2025challenge}. 
Still, it remains an ongoing field of research to determine an optimal interface between the foundation model and the robot.
Some methods sequence learned policies via natural language, effectively using LLMs as a drop-in replacement for a task planner \cite{ahn2022can, driess23a, huang2023inner, lin2023text2motion}. 
Other works prompt or fine-tune LLMs to directly predict low-level control signals \cite{wang2024prompt, zhu2024incoro, zitkovich2023rt, brohan2022rt}.
However, natural language lacks the required geometric precision for many tasks, and numerical control outputs are difficult to predict for LLMs.
Therefore, code has emerged as a promising paradigm to link foundation model and control policy.
Examples include reward function optimization using LLMs \cite{ma2024eureka, yu2023language, song2023self}, writing robot policy code directly \cite{liang2023code, burns2024genchip, mu2024robocodex, hu2024deploying}, and proposing constraints for trajectory optimization which can be used to generate trajectories \cite{huang2023voxposer, curtis2024trust}.
Our method follows this last approach, as we also use a foundation model to optimize trajectory constraints.
In difference to prior work like PRoC3S \cite{curtis2024trust}, however, we optimize the proposed constraints with a zero-order numerical optimization method to further improve the geometric accuracy of the optimization problem.

\paragraph{Foundation Models for Optimization.}
Large pretrained neural networks are commonly referred to as foundation models (FMs) \cite{bommasani2021opportunities}.
While most models are still commonly referred to as \textit{language} models, recent work has demonstrated that FMs can serve as versatile optimizers in various contexts.
Examples of this include regression or sequence completion \cite{mirchandani2023largelanguagemodelsgeneral}, code optimization \cite{lange2025ai, morris2024llm}, numerical optimization \cite{lange2024large}, as well as mathematical program optimization \cite{romera2024mathematical} and problem solving \cite{trinh2024solving}.
At the heart of these FM-based optimization approaches lies iterative prompting and an external fitness or cost function that quantifies the quality of a solution.
These ingredients establish a strong connection between evolutionary computation and FMs \cite{song2024position}, as both frameworks continuously modify a set of solution candidates to improve their expected performance.
Our work follows this framework, but applies within the field of robotics, as we use an LLM to continuously improve a trajectory optimization problem that solves a specified task.

%% file: content/experiments.tex
\begin{figure}[t]
  \centering
  \includegraphics[page=2, width=.8\textwidth, trim={10cm 13cm 10cm 13cm}, clip]{imgs/mops_plots.pdf}
  \subfloat[Code as Policies]{%
    \includegraphics[width=0.28\textwidth, trim={0cm 0.5cm 2cm 0cm}, clip]{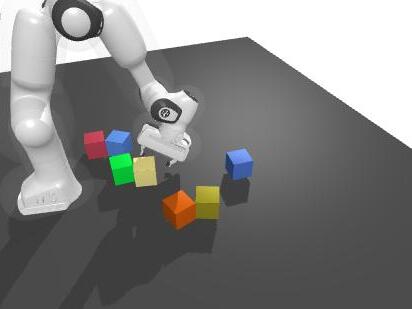}%
  }\hfill
  \subfloat[PRoC3S]{%
    \includegraphics[width=0.28\textwidth, trim={0cm 0.5cm 2cm 0cm}, clip]{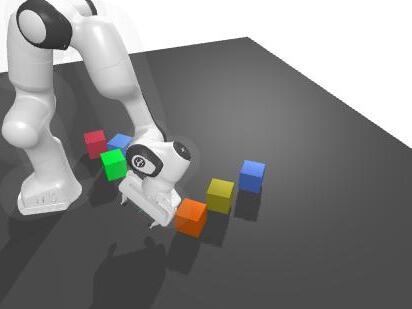}%
  }\hfill
  \subfloat[MOPS (Ours)]{%
    \includegraphics[width=0.28\textwidth, trim={0cm 0.5cm 2cm 0cm}, clip]{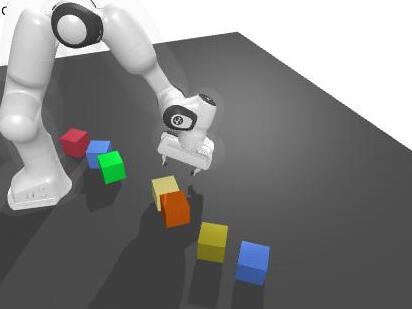}%
  }
  \caption{Solutions produced by all evaluated methods in the `Pushing' domain. MOPS is the only method that achieves a straight line that includes all blocks.}
  \label{fig:qual_res}
\end{figure}

The goal of this section is to answer three central questions: \textit{(i)} how does MOPS compare to prior language-conditioned TAMP methods?; \textit{(ii)} what role does the black-box optimizer play in our approach?; \textit{(iii)} how important is the LLM feedback loop?

\subsection{Experimental Setup}

\paragraph{Task Environments.} We evaluate our method across two distinct domains. 
In the \emph{Pushing} environment, a Panda robot must push multiple cuboids into predefined goal configurations.
This environment features three specific tasks: \textit{(i)} arranging blocks in a circle, \textit{(ii)} arranging blocks in a straight line, and \textit{(iii)} maneuvering a block around a wall to a target position. 
The integration of static and dynamic obstacles—such as a wall and other movable blocks—significantly elevates the planning complexity by constraining feasible trajectories, thereby requiring the development of more sophisticated manipulation strategies that account for environmental constraints.

In the \emph{Drawing} environment, a Panda robot must draw on a whiteboard of fixed dimensions but variable tilt angle. 
We adapted this environment from \citet{curtis2024trust}, but made it considerably more challenging by introducing a tilt to the whiteboard.
A top-down camera, aligned with the table frame, captures visual observations of the scene. 
The objective is to produce symbols that appear visually accurate despite the perspective distortion induced by the angular mismatch between the camera and whiteboard reference frames. 
We evaluate performance on three specific drawing tasks: (1) a five-pointed star, (2) a regular pentagon, and (3) an the hash character \#.

\paragraph{Baseline Methods.} 
We compare against two leading language-conditioned TAMP approaches.
As discussed above, a common line of work on language-conditioned TAMP uses the FM to directly generate executable program code.
Among these methods, we adopt Code-as-Policies (CaP) \cite{liang2023code} as a state-of-the-art representative.
CaP leverages FMs to generate complete policy code, including both high-level decision logic and auxiliary helper functions that interface with predefined Python functions for perception, planning, and control. 
An alternative is to optimize the generated plan in a closed-loop manner, as we do. 
The closest method in this vein is PRoC3S \cite{curtis2024trust}, which closes the plan-generation loop but uses a uniform sampler to produce continuous action parameters. 
In contrast, MOPS searches over sequences of parameterized \emph{constraints} and tunes their parameters via black-box optimization rather than random sampling. 
Although we aggregate multiple constraints into compact building blocks to reduce dimensionality in our experiments, MOPS can reason at the level of individual constraints in principle. 
This stands in stark contrast to approaches that require action abstractions for planning.
To evaluate the efficacy of these improvements, we therefore adopt PRoC3S as an additional baseline method.

\begin{figure}[t]
    \centering
    \subfloat[Pushing: Circle]{\includegraphics[width=0.3\textwidth]{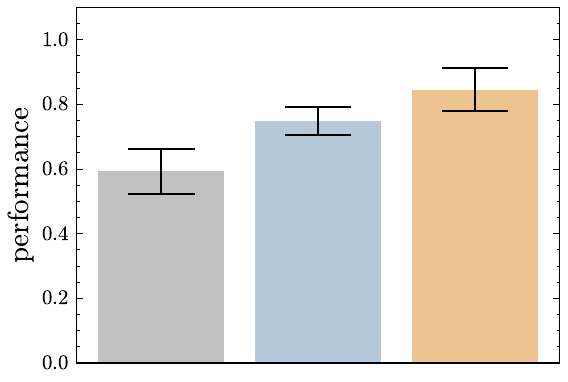}} \hfill
    \subfloat[Pushing: Line]{\includegraphics[width=0.3\textwidth]{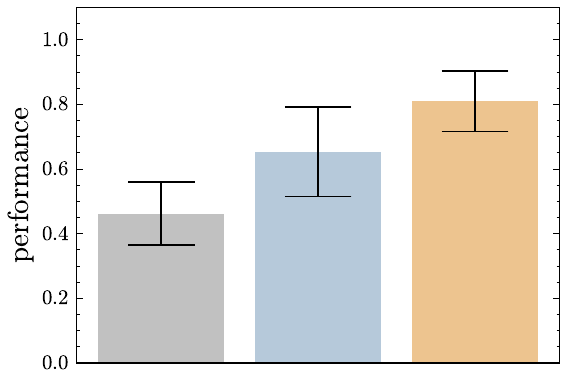}} \hfill
    \subfloat[Pushing: Obstacle Avoidance]{\includegraphics[width=0.32\textwidth]{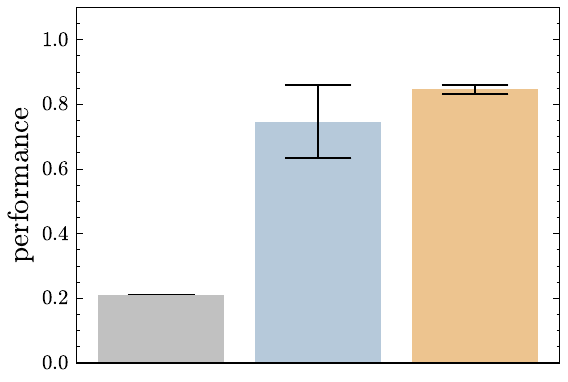}}\\ 
    
    \subfloat[Drawing: Star]{\includegraphics[width=0.3\textwidth]{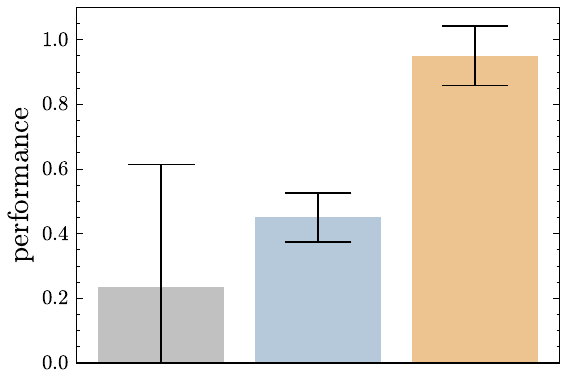}} \hfill
    \subfloat[Drawing: Pentagon]{\includegraphics[width=0.3\textwidth]{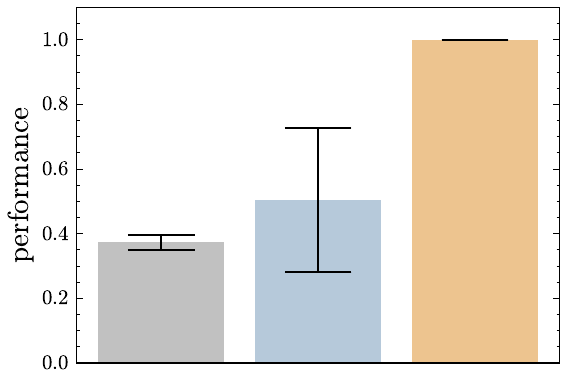}} \hfill
    \subfloat[Drawing: Hash]{\includegraphics[width=0.3\textwidth]{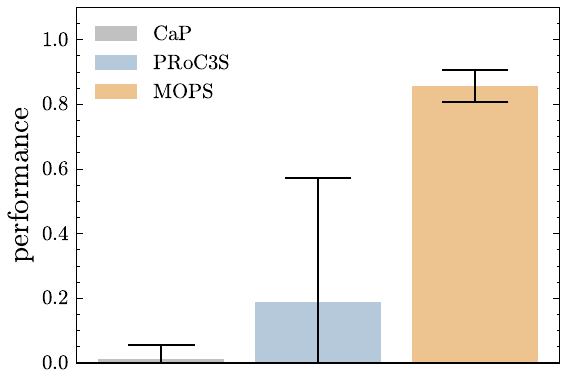}}
    \caption{
    Normalized performances across six challenging tasks. MOPS outperforms the baselines across all tasks. We report averaged normalized performances across 5 independent seeds ($\pm 1.96$ standard deviations). }\label{fig:res_comparison}
    \label{fig:experim_results}
\end{figure}


\subsection{Analyzing Performance}\label{secExp} 
To answer how MOPS compares to prior work, we evaluate MOPS and all baselines on all presented tasks.
For PRoC3S and our approach, we allocate $1{,}000$ sampling/optimization steps per LLM query in the drawing domain and $1{,}500$ in the block pushing domain due to its more challenging nature. 
A maximum of 2 feedback iterations were permitted, limiting the total FM prompts to 3.
Further experimental details are elaborated in the Appendix.
The results of the experiments are depicted in Fig.\ \ref{fig:experim_results}, and an aggregation across domains can be found in Fig.\ \ref{fig:visualAbstract}.
We observe that our method outperforms the baselines on all tasks, or is at least as good as them.
In the challenging drawing domain, it is apparent that optimization-free approaches such as CaP cannot solve any tasks, as the initial guess of the FM for the action sequences and precise motions lacks precision.
Further, we observe that our method improves over sampling-based prior work (PRoC3S) by introducing an optimization step of the constraint parameters.
This is particularly visible in Fig.\ \ref{fig:qual_res} which displays the qualitative results in the pushing domain.
Due to the number of blocks in this domain, the search space dimension for numerical parameters is too high for methods that employ random sampling.

\begin{figure}[t]
    \centering
    \subfloat[Pushing: Circle]{\includegraphics[width=0.3\textwidth]{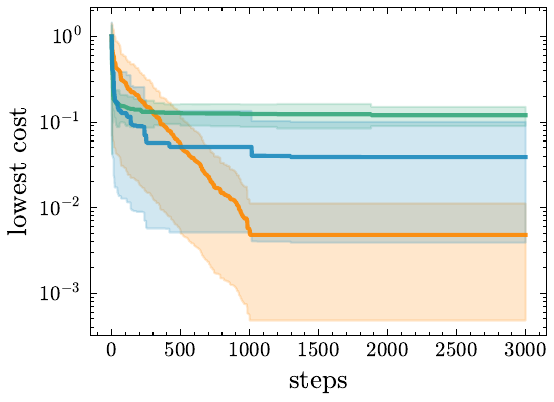}} \hfill
    \subfloat[Drawing: Pentagon]{\includegraphics[width=0.3\textwidth]{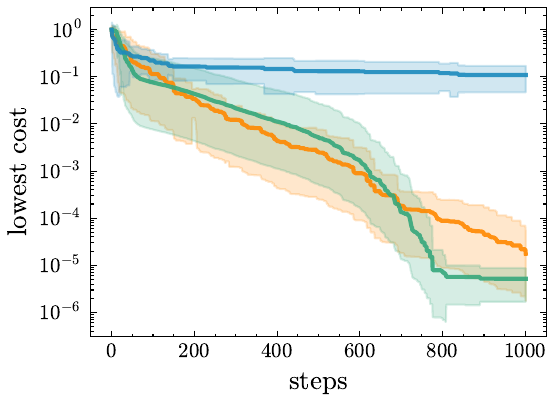}} \hfill
    \subfloat[Drawing: Star]{\includegraphics[width=0.3\textwidth]{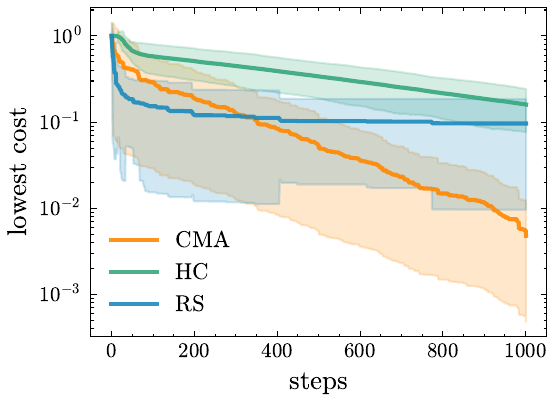}}
    \caption{Results comparing different BBO methods for constraint parameter optimization: CMA-ES, probabilistic hill climbing (HC), and random sampling (RS).
    We observe that CMA-ES performs the best across tasks. 
    Random sampling can be efficient if the initial guess is good, but fails if this is not the case.
    We report averaged normalized performances across 5 independent seeds ($\pm 1.96$ standard deviations). 
    In addition, we report further results across all tasks are provided in the Appendix~\ref{fig:ablations_app}.
    }
    \label{fig:ablations_main}
\end{figure}

\begin{wrapfigure}{R}{0.35\textwidth}
  \begin{center}
    \includegraphics[width=\linewidth]{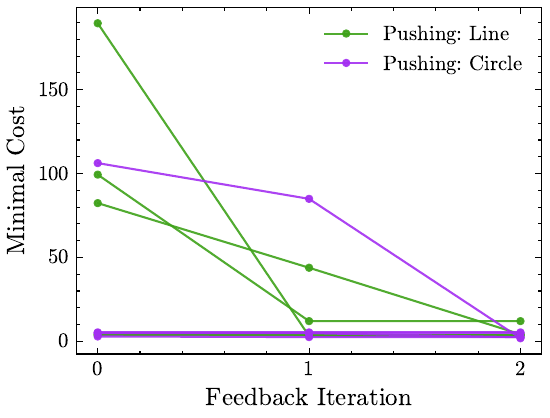}
  \end{center}
  \caption{Cost over feedback iterations across 5 independent seeds per task.}
  \label{fig:feedback_cost}
  \label{fig:fb}
\end{wrapfigure}

\subsection{What matters in MOPS?}
To understand the contribution of individual components within our proposed method, we conduct an ablation study that evaluates the continuous parameter optimization. 
Specifically, we systematically investigated the role of the inner blackbox optimization loop by comparing three distinct optimization strategies: random sampling (as in PRoC3S), CMA-ES \cite{hansen2003reducing}, and a probabilistic variant of hill climbing that explores parameter space through directional perturbations \cite{romeo1984probabilistic}. 
Results for three representative tasks are shown in Figure~\ref{fig:ablations_main}.
The experiments demonstrate that optimization clearly improves the initial guess from the LLM.
While random sampling performs well when the initial guess from the LLM is of high quality, it falls short of achieving substantial improvements on the more challenging tasks. In contrast, CMA-ES rapidly identifies the appropriate constraint parameters to minimize the task cost.
Further, we analyze the role of the outer feedback loop in Fig.~\ref{fig:fb}.
We see that LLM feedback is crucial for reliable problem solving. 
While the initial constraint set is sufficient on some runs, the variance across runs is high when no feedback is used and low after 2 rounds of feedback.
We expand this analysis in Appendix \ref{secFeed}.


%% file: content/conclusion.tex
In this work, we present Meta-Optimization and Program Search (MOPS), a method for TAMP that optimizes sequences of constraints that induce motions to satisfy a language instructed goal. 
In contrast to prior work, MOPS meta-optimizes trajectory optimization programs using a mix of LLM-, and numerical black-box optimization, enabling it to solve complex manipulation tasks.
We conduct a comprehensive evaluation across multiple tasks in diverse environments.
Our results show that MOPS improves over prior TAMP approaches.

%% file: content/limitations.tex
While our method offers flexibility and performance across different robotic tasks, it comes with certain limitations. 
First, it requires the tuning of additional hyperparameters introduced by the inner optimizer (e.g., the step size in hill climbing or the initial sigma in CMA-ES), which may affect robustness and reproducibility. 
This issue may be mitigated by making part of the hyperparameters tunable by the foundation model, similar to the initial guess prediction that is already part of the method.
Second, our approach assumes the availability of a task-specific cost function $\Psi$, which must be designed to reflect the desired task outcomes. 
Future work could explore methods for learning or predicting cost functions to improve generality.
Further, our current experiments optimize over functions that aggregate multiple constraints due to the limitations of VLMs at the time. We believe that future model releases will permit to reason directly at the individual constraint level.
Lastly, our method relies on full state knowledge of the scene; combining with VLMs or incorporating state estimation techniques could help relax this requirement.



%% file: content/appendix.tex
\appendix

\section{Additional Results}
\subsection{Optimizer Ablation Study}
\begin{figure}[htb]
    \centering
    \subfloat[Pushing: Circle]{\includegraphics[width=0.3\textwidth]{imgs/circle_cost_comparison.pdf}} \hfill
    \subfloat[Drawing: Pentagon]{\includegraphics[width=0.3\textwidth]{imgs/pentagon_cost_comparison.pdf}} \hfill
    \subfloat[Drawing: Star]{\includegraphics[width=0.3\textwidth]{imgs/star_cost_comparison.pdf}} \\
    \subfloat[Pushing: Line]{\includegraphics[width=0.3\textwidth]{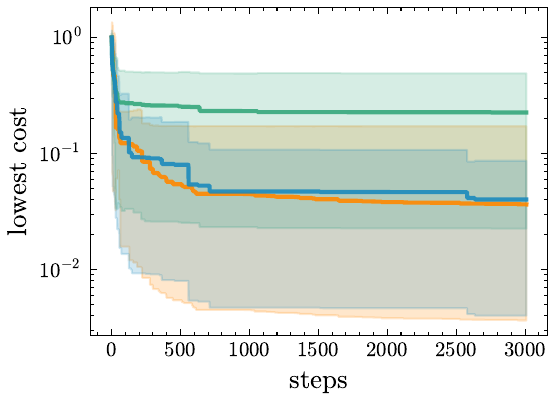}} \hfill
    \subfloat[Pushing: Obstacle Avoidance]{\includegraphics[width=0.3\textwidth]{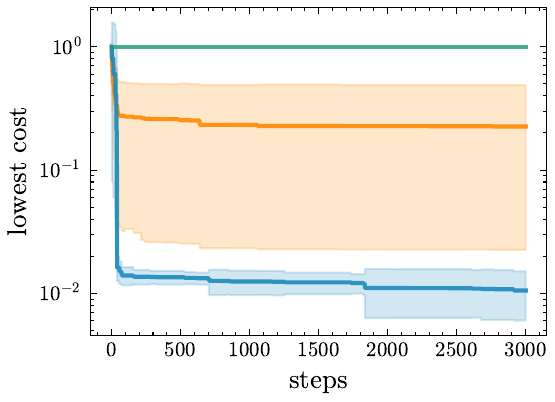}} \hfill
    \subfloat[Drawing: Hash]{\includegraphics[width=0.3\textwidth]{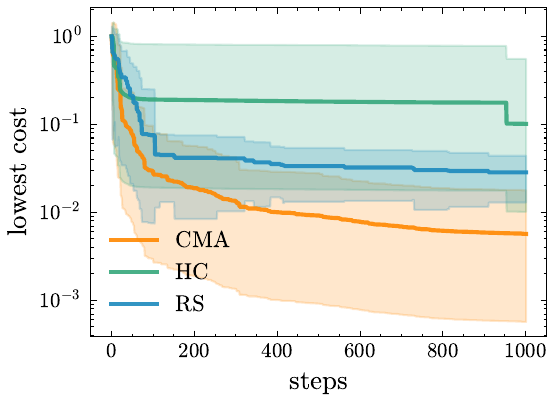}}
    \caption{Complete ablation study comparing different BBO methods for constraint parameter optimization across all six simulated manipulation tasks. We evaluate CMA-ES, probabilistic hill climbing (HC), and random sampling (RS) on both pushing and drawing tasks.  We report averaged normalized performances across 5 independent seeds ($\pm 1.96$ standard deviations). The top row shows the main results discussed in Section~\ref{sec:Experiments}, while the bottom row provides additional task variations that confirm the observed performance trends.}
    \label{fig:ablations_complete}
\end{figure}

For completeness, we list the full BBO experiment results across all tasks in Fig.\ \ref{fig:ablations_complete}.
These results complement Fig.\ \ref{fig:ablations_main}, which is truncated in the main part owing to space constraints.
The full results confirm the results from the main paper. 
Across four of the six tasks, CMA-ES performs the best. For the pentagon drawing task, hill climbing performs the best. Random sampling performed best for this iteration of the obstacle avoidance task.

\subsection{Qualitative Results for Drawing Tasks}
\begin{figure}[htb]
    \centering
    \subfloat[CaP]{\includegraphics[width=0.3\textwidth]{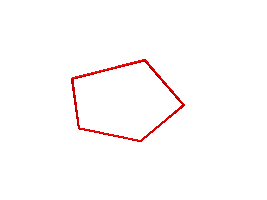}}\hfill
    \subfloat[PRoC3S]{\includegraphics[width=0.3\textwidth]{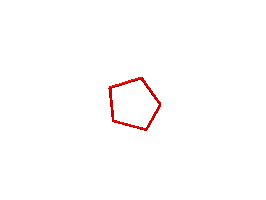}}\hfill
    \subfloat[MOPS]{\includegraphics[width=0.3\textwidth]{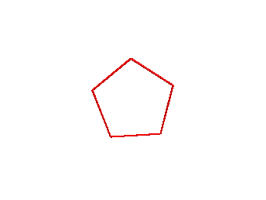}} \\[1ex]
    \subfloat[CaP]{\includegraphics[width=0.3\textwidth]{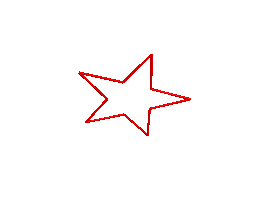}}\hfill
    \subfloat[PRoC3S]{\includegraphics[width=0.3\textwidth]{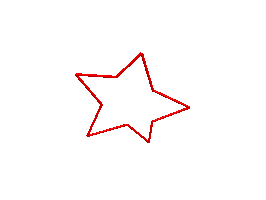}}\hfill
    \subfloat[MOPS]{\includegraphics[width=0.3\textwidth]{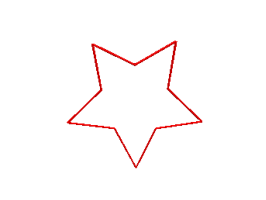}} \\[1ex]
    \subfloat[CaP]{\includegraphics[width=0.3\textwidth]{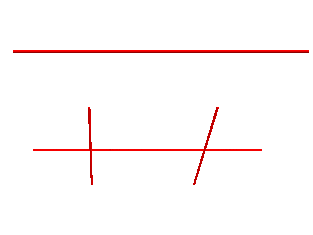}}\hfill
    \subfloat[PRoC3S]{\includegraphics[width=0.3\textwidth]{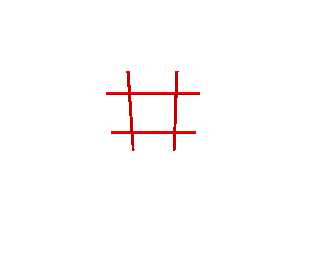}}\hfill
    \subfloat[MOPS]{\includegraphics[width=0.3\textwidth]{imgs/hash_mops.png}}
    \caption{Resulting images across methods in the drawing environment.}
    \label{fig:ablations_app}
\end{figure}

For completeness, we list the qualitative results on the drawing task.
That is, we provide the final solutions in Fig. \ref{fig:ablations_app}.
The drawing task evaluates the system's ability to produce visually accurate symbols despite perspective distortion caused by the angular mismatch between the whiteboard tilt angle and the camera's reference frame. 
The quantitative results in \ref{fig:experim_results} demonstrate that MOPS performs the best on the drawing tasks.
The qualitative results illustrate this.
The baseline methods, even when provided with complete state information (i.e. camera intrinsics, extrinsics, and global whiteboard position), struggle to produce accurate drawings in the resulting images, highlighting the inherent complexity of this challenge.
Our proposed method MOPS significantly outperforms baseline methods by exploiting gradient information within the cost function to optimize line drawing parameters for perceptual accuracy in the image space. 
By accounting for perspective effects, MOPS generates drawings that look accurate when viewed through the camera.

\newpage
\subsection{Detailed Feedback Loop Study}\label{secFeed}
We analyze the contribution of the outer feedback loop, i.e. the language model program search to our method's performance. In the \emph{Drawing} environment, we observed that the outer feedback loop did not significantly improve performance: the FM typically predicted the correct number of lines on the first attempt, after which the inner optimizer converged successfully. Nevertheless, the feedback loop becomes crucial in scenarios where the FM fails to provide a good enough zero-shot guess (e.g., producing an incorrect number of lines).

The importance of the feedback loop becomes clear in the \emph{Pushing} environment. Experimental results, performed across 5 runs per task, highlight its importance for achieving the reported performance (Fig.~\ref{fig:feedback_cost}).

\newpage
\subsection{Real-world experiments}

\begin{wrapfigure}{r}{0.5\textwidth}
  \vspace{-.5 in}
  \begin{center}
    \subfloat[Drawing]{\includegraphics[height=0.25\textwidth]{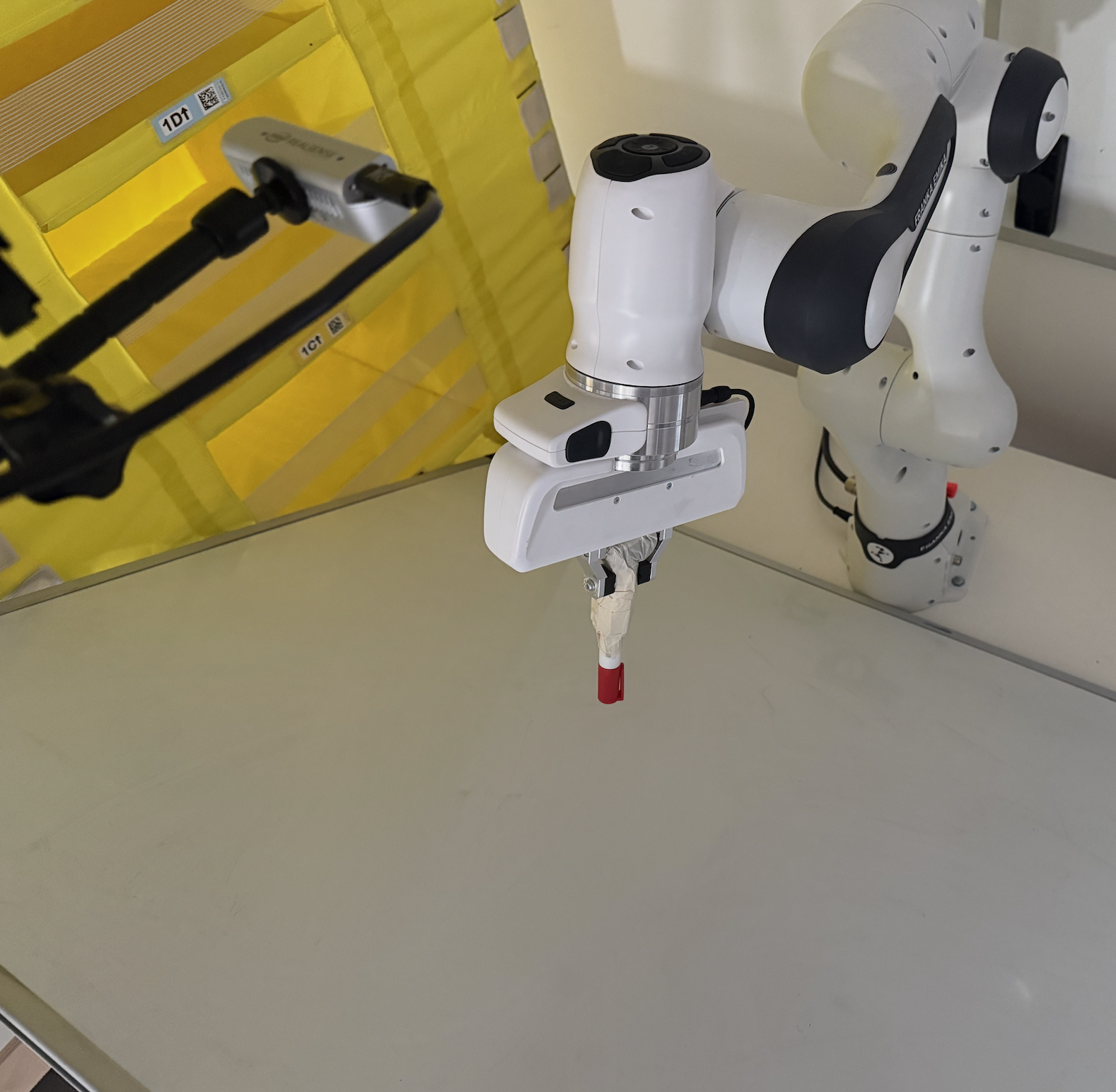}}%
    \hspace{0.05\textwidth}%
    \subfloat[Block pushing]{\includegraphics[height=0.25\textwidth]{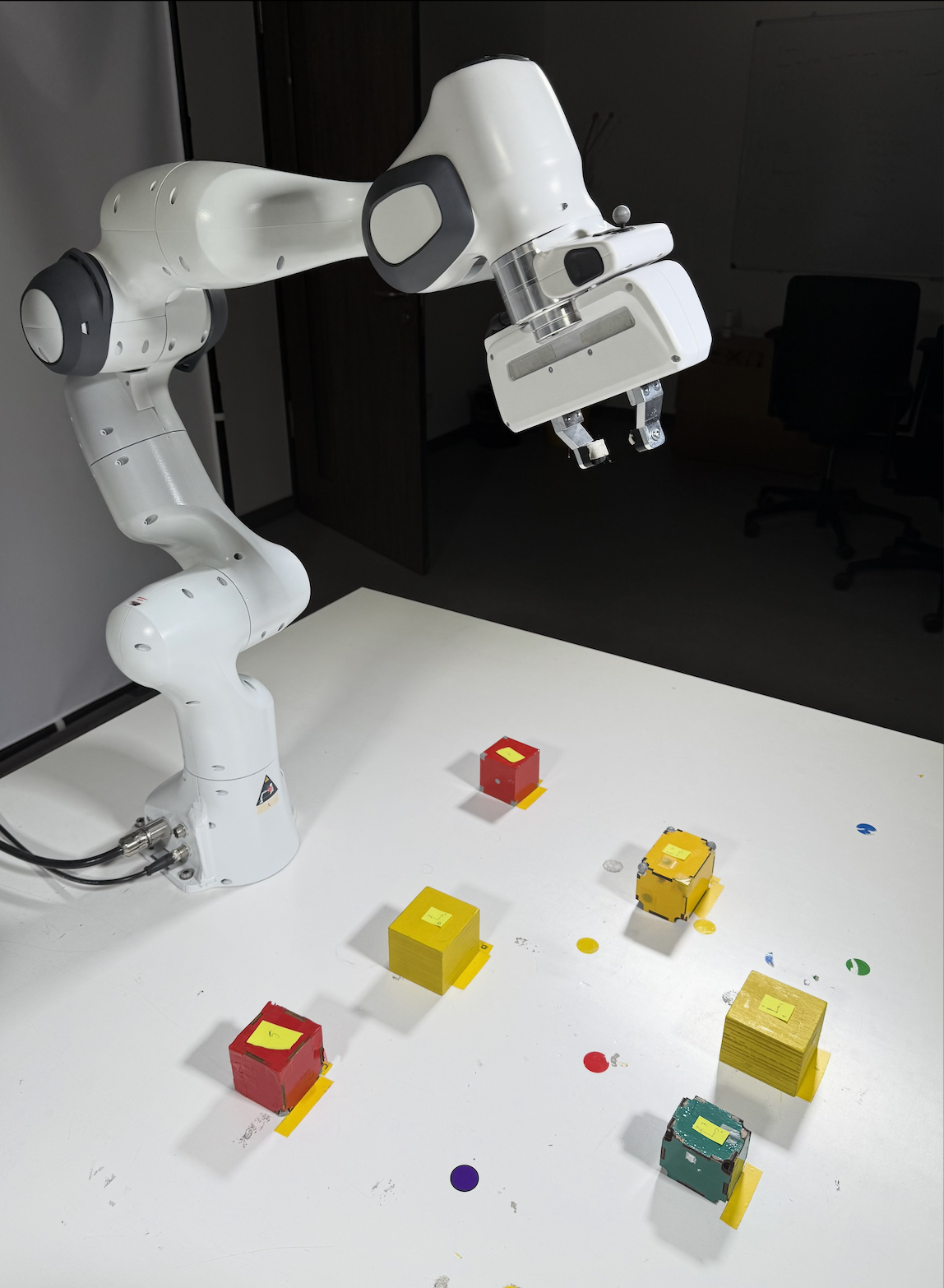}}%
  \end{center}
  \vspace{-0.1in}
  \caption{Real-world experiment setup.}
  \label{fig:real}
  \vspace{-0.1in}
\end{wrapfigure}

To assess the real-world performance of our approach, we deploy it on the Franka Panda robot platform. 
The experimental setups are shown in Figure~\ref{fig:real}. 
While our method transfers successfully to the real robot, careful calibration of the physical setup is required, as simulation provides full state information that is not directly accessible in the real world. 
Qualitative results and successful executions can be found in the supplementary video.

\subsection{LLM Ablation Study}

We evaluated a wide range of LLMs and VLMs on code generation for our method. For each test, we examined the outputs produced by the models and assessed whether they correctly followed the instructions in the initial prompt. Results were averaged over five runs. The results are shown in Table \ref{tab:LLMab}. The most common failure modes were \textit{(i)} hallucinating new high-level functions not specified in the prompt, and \textit{(ii)} omitting key instructions, such as importing required libraries outside the generated functions or creating extra functions when explicitly instructed not to. Performance improvements for smaller models through fine-tuning may be possible; we leave this exploration to future work.

\begin{table}[h]
  \centering
  \begin{tabular}{|c|c|}
    \hline
    \textbf{Model} & \textbf{Success Rate} \\ \hline
    Qwen2-VL-7B-Instruct & 0\% \\ \hline
    Qwen2.5-14B-Instruct-1M & 0\% \\ \hline
    Qwen2.5-Coder-32B-Instruct & 0\% \\ \hline
    Qwen2.5-VL-72B-Instruct & 0\% \\ \hline
    DeepSeek-R1 & 100\% \\ \hline
    GPT-3.5-turbo & 60\% \\ \hline
    GPT-4-turbo & 100\% \\ \hline
    GPT-4o-mini & 100\% \\ \hline
  \end{tabular}
  \vspace{0.5em}
  \caption{Comparison between different LLMs and VLMs.}
  \label{tab:LLMab}
\end{table}

\section{Experimental Details}
This section lists the full experimental details for this paper.

The code is partially based on \citet{curtis2024trust}.
For the experiments, we use the baseline implementations of CaP and PRoC3S of this repository.
For CMA-ES, we use the Python implementation of pycma \cite{hansen2022cma}.
All mathematical programs are implemented and optimized via our lab's trajectory optimization library.
All experiments we performed on an internal cluster with 12-core CPUs and 32GB of RAM.
The code to reproduce our experiments and plots will be made available upon conference publication.
For each method we used the OpenAI GPT-4o FM \cite{hurst2024gpt}, specifically the gpt-4o-mini-2024-07-18 checkpoint.
To evaluate the task cost, i.e., success, we simulate the trajectories in the NVIDIA PhysX simulator \cite{nvidia-physx}.

Each experiment is repeated 5 times using randomly generated seeds.
In each plot, we report the mean performances across all 5 runs and 1.96 standard deviations.
To align the cost function scores across tasks, performance is reported as 1 - log-normalized error, calculated by applying $\log(1+x)$ transformation to raw error values and normalizing by the maximum possible error. 
Standard deviations undergo identical transformation to preserve scaling consistency. 
This metric provides a more interpretable performance assessment where higher values indicate better performance, with 1.0 representing perfect execution. 

For PRoC3s and our approach, we allocated 1000 sampling/optimization steps per trial in the drawing domain and 1500 in the block pushing domain. 
A maximum of 2 feedback iterations were permitted, limiting the total FM prompts to 3.
For the BBO ablation, we slightly deviate from the general setup and increase the number of independent runs to 10 in the draw environment. 
To isolate the effect of the optimization method, the initial plan generated by the LLM was fixed across all experimental configurations.

The initial sigma value for the CMA-ES optimizer differs between the two task domains: it is set to 0.01 for the drawing task and 0.05 for the pushing task.

\section{Cost Function Details}

This section provides the cost functions for each of the tasks in both the pushing and drawing domains.

\subsection{\emph{Pushing} environment.}

\textbf{Push Box and Avoid Obstacle:} Let $p_{\text{box}}$ be the position of the box, $p_{\text{target}}$ the target position, $p_{\text{grip}}$ the gripper position, 
$p_{\text{wall}}$ the wall position, and $p_{\text{init}}$ the initial box position. 
The individual costs are

\begin{align}
C_{\text{pos}}         &= \Bigl( 4 \cdot ( p_{\text{box}} - p_{\text{target}} ) \Bigr)^2 \\
C_{\text{wall}}        &= -\log \Bigl( \| p_{\text{wall}}^{(xy)} - p_{\text{box}}^{(xy)} \| \Bigr) \\
C_{\text{init}}        &= -\log \Bigl( \max \{ \| p_{\text{box}} - p_{\text{init}} \|, 0.001 \} \Bigr) \\
C_{\text{endeff}}      &= \Bigl( 4 \cdot ( p_{\text{box}} - p_{\text{grip}} ) \Bigr)^2 \\
C_{\text{endeff-wall}} &= - \| p_{\text{wall}} - p_{\text{grip}} \|
\end{align}

The total cost is given by
\begin{equation}
C_{\text{total}} =
2.0 \, C_{\text{pos}} + 
0.01 \, C_{\text{wall}} +
0.01 \, C_{\text{init}} +
0.7 \, C_{\text{endeff}} +
0.2 \, C_{\text{endeff-wall}}.
\end{equation}

\textbf{Push Circle:} Given points $P = \{p_1, \ldots, p_n\}$ with center $c = \frac{1}{n}\sum_{i=1}^{n} p_i$, we minimize $C_{\text{total}} = 1000 \cdot C_{\text{rad}} + C_{\text{neigh}}$ where $C_{\text{rad}} = \sum_{i=1}^{n} (0.2 - \|p_i - c\|)^2$ and $C_{\text{neigh}} = \sum_{i=1}^{n} (0.2 - d_i)^2$ with $d_i = \min_{j \neq i} \|p_j - p_i\|$.

\textbf{Push Line:} Given points $P = \{p_1, \ldots, p_n\}$ where $p_i = (x_i, y_i)$, we minimize $C_{\text{total}} = 10^4 \cdot C_{\text{fit}} + 10^2 \cdot C_{\text{space}}$ where $C_{\text{fit}} = \frac{1}{n} \sum_{i=1}^{n} (y_i - (\hat{m}x_i + \hat{b}))^2$ for best-fit line $L: y = \hat{m}x + \hat{b}$, and $C_{\text{space}} = \frac{1}{n-1}\sum_{k=1}^{n-1} (d_k - \bar{d})^2$ with $d_k = \|p'_{(k+1)} - p'_{(k)}\|$ for consecutive projected points on $L$ and $\bar{d} = \frac{1}{n-1}\sum_{k=1}^{n-1} d_k$.

\subsection{\emph{Drawing} environment.}
The drawing domain involves more rigorous and fine-tuned cost functions, as it needs to account for connectivity, length, angles, spacings, inner/outer radii, parallelism and further edge cases. 
\\
\textbf{Draw Pentagon:}
$
C_{\text{total}} = C_{\text{conn}} + C_{\text{len}} + C_{\text{rad}} + C_{\text{angle}} + C_{\text{spacing}} + C_{\text{close}} + P_{\text{size}}$
or, together:
\begin{align*}
C_{\text{total}} &= 
500 \sum_{i=0}^{4} \| (s_i + v_i) - s_{(i+1) \bmod 5} \|^2
+ 500 \frac{\mathrm{Var}(\|v\|)}{\left( \frac{1}{5} \sum_{i=0}^4 \|v_i\| \right)^2} \\
&\quad+ 300 \frac{\mathrm{Var}(\|s_i - c\|)}{\left( \frac{1}{5} \sum_{i=0}^4 \|s_i - c\| \right)^2}, \quad c = \frac{1}{5}\sum_{j=0}^{4} s_j \\
&\quad+ 100 \sum_{i=0}^4 \left[ \arccos\!\left( \frac{-v_{(i-1) \bmod 5} \cdot v_i}{\|v_{(i-1) \bmod 5}\|\, \|v_i\|} \right) - \frac{3\pi}{5} \right]^2 \\
&\quad+ 100 \sum_{i=0}^4 \left[ \arccos\!\left( \frac{(s_i - c) \cdot (s_{(i+1) \bmod 5} - c)}{\|s_i - c\|\, \|s_{(i+1) \bmod 5} - c\|} \right) - \frac{2\pi}{5} \right]^2 \\
&\quad+ 200 \,\| s_0 - (s_4 + v_4) \|^2
+ \mathbf{1}\!\left[ \frac{1}{5}\sum_{i=0}^4 \|v_i\| < L_{\text{thresh}} \right] \cdot 700
\end{align*}

\textbf{Draw Star:}
$
C_{\text{total}} = C_{\text{conn}} + C_{\text{rad,outer}} + C_{\text{rad,inner}} + C_{\text{ratio}} + C_{\text{angle}} + P_{\text{size}}
$
or, together:
\begin{align*}
C_{\text{total}} &= 
100 \sum_{i=0}^{9} \| (s_i + v_i) - s_{(i+1) \bmod 10} \|^2 \\
&\quad+ 500 \frac{\mathrm{Var}\big( \{\|s_i - c\| : i \in I_{\text{outer}}\} \big)}{(\bar{r}_{\text{outer}})^2}
+ 500 \frac{\mathrm{Var}\big( \{\|s_i - c\| : i \in I_{\text{inner}}\} \big)}{(\bar{r}_{\text{inner}})^2} \\
&\quad+ \begin{cases}
300(1.5 - \rho)^2, & \rho < 1.5 \\
100(\rho - 2.0)^2, & \text{otherwise}
\end{cases} \quad \rho = \frac{\bar{r}_{\text{outer}}}{\bar{r}_{\text{inner}}} \\
&\quad+ 100 \sum_{i=0}^9 \left[ \arccos\!\left( \frac{(s_i - c) \cdot (s_{(i+1) \bmod 10} - c)}{\|s_i - c\|\, \|s_{(i+1) \bmod 10} - c\|} \right) - \frac{\pi}{5} \right]^2 \\
&\quad+ \mathbf{1}\!\left[ \frac{1}{10}\sum_{i=0}^9 \|v_i\| < 35 \right] \cdot 700
\end{align*}
where:

\begin{align*}
c &= \frac{1}{10}\sum_{j=0}^{9} s_j, \quad 
I_{\text{outer}} = \{0, 2, 4, 6, 8\}, \quad I_{\text{inner}} = \{1, 3, 5, 7, 9\}, \\
\bar{r}_{\text{outer}} &= \frac{1}{5}\sum_{k \in I_{\text{outer}}} \|s_k - c\|, \quad
\bar{r}_{\text{inner}} = \frac{1}{5}\sum_{k \in I_{\text{inner}}} \|s_k - c\|
\end{align*}

\textbf{Draw hash:}
$
C_{\text{total}} = C_{\text{straight}} + C_{\text{parallel}} + C_{\text{perp}} + C_{\text{spacing}} + C_{\text{intersect}} + C_{\text{len\_con}} + C_{\text{len\_bal}} + P_{\text{size}}
$
or, together:
\begin{align*}
C_{\text{total}} &= 
100 \!\left( \sum_{v \in V_h} \left(\frac{|v_y|}{\|v\|}\right)^2 + \sum_{v \in V_v} \left(\frac{|v_x|}{\|v\|}\right)^2 \right) \\
&\quad+ 50 \big( (\Delta\alpha_h)^2 + (\Delta\alpha_v)^2 \big) \\
&\quad+ 50 \sum_{v_h \in V_h} \sum_{v_v \in V_v} \left( \frac{v_h \cdot v_v}{\|v_h\| \, \|v_v\|} \right)^2 \\
&\quad+ 300\big((d_h - 0.33)^2 + (d_v - 0.33)^2\big) + 100(d_h - d_v)^2 \\
&\quad+ \sum_{v_h \in V_h} \sum_{v_v \in V_v} \mathrm{Penalty}_{\text{int}}(v_h, v_v) \\
&\quad+ 50\!\left( \left(\frac{\|v_{h1}\| - \|v_{h2}\|}{\frac{1}{2}(\|v_{h1}\|+\|v_{h2}\|)}\right)^2 
+ \left(\frac{\|v_{v1}\| - \|v_{v2}\|}{\frac{1}{2}(\|v_{v1}\|+\|v_{v2}\|)}\right)^2 \right) \\
&\quad+ 20 \left( \frac{\sum_{v \in V_h} \|v\| - \sum_{v \in V_v} \|v\|}{\sum_{v \in V} \|v\|} \right)^2 \\
&\quad+ \mathbf{1}\!\left[ \bar{l} < 75 \right] \cdot 700
\end{align*}
where:

\begin{align*}
&V_h = \{v \in V \mid |v_x| > |v_y|\}, \quad V_v = \{v \in V \mid |v_x| \le |v_y|\}, \\
&S_h, S_v \text{ are the corresponding starting points}, \\
&d_h = \frac{|s_{h1,y} - s_{h2,y}|}{\bar{l}}, \quad d_v = \frac{|s_{v1,x} - s_{v2,x}|}{\bar{l}}, \quad 
\bar{l} = \frac{1}{4} \sum_{i=0}^3 \|v_i\|, \\
&\Delta\alpha_h, \Delta\alpha_v \text{ are the minimum angle differences between parallel vectors}, \\
&\mathrm{Penalty}_{\text{int}} \text{ is high if segments fail to intersect near } \frac{1}{3} \text{ or } \frac{2}{3} \text{ along each segment}.
\end{align*}

\section{MOPS Prompting Details}\label{secPrompts}
\input{content/prompts}

\clearpage

%% file: content/prompts.tex
In this section, we present the prompts used in our method, which follow the general structure introduced in PRoC3S \cite{curtis2024trust}.

\subsection{Prompt for Draw Environment}

\begin{lstlisting}

You are a franka panda robot operating in an environment with the following state:

The whiteboard is bounded in the x-direction between 0 and 0.64, and in the y-direction between 0 and 0.48.
The whiteboards world position is at (0.0, 0.4, 0.96).
The whiteboard is tilted by 40 degrees.
A camera, positioned at (0.0, 0.45, 1.5), looks downward parallel to the world x-y plane.

You have access to the following set of skills expressed as pddl predicates followed by descriptions. 
You have no other skills you can use, and you must exactly follow the number of inputs described below.
The coordinate system is defined relative to the whiteboard, using x and y axes. The x-axis runs horizontally along the whiteboard, while the y-axis runs vertically on it. The origin (0, 0) is located at the lower-left corner of the whiteboard.

Action("draw_line", [x0, y0, x1, y1])
Draw a line on a Whiteboard, x0, y0 being the start point of the line on the whiteboard coordinates, x1, y1 the endpoints.

Additionally, the input to `gen_initial_guess` must be exactly the `initial:DrawState` argument, even if this isn't explicitly used within the function!
The `gen_initial_guess` function MUST return a dict.
If you need to import any modules DO IT INSIDE the `gen_plan` function.
To compensate for an uneven drawing surface and ensure a flat appearance in the top-down camera view, apply 2D offsets to each point, initialized to [0., 0.].
ALWAYS ADD POINT OFFSETS INITIALIZED TO ZERO TO THE `gen_initial_guess`!

Below is one example for a tasks and successful solutions.

# user message
State: DrawState(frames=[])
Goal: Draw a square on the tilted Whiteboard with side lengths of 20cm. 

# assistant message
```python
def gen_plan(state: DrawState, pos: list, size: float, offsets: list):
    x, y = pos

    # Define base square corners in order (clockwise)
    base_corners = [
        [x, y],                   # bottom-left
        [x + size, y],            # bottom-right
        [x + size, y + size],     # top-right
        [x, y + size]             # top-left
    ]

    # Add offsets to each corner
    perturbed = [
        [px + dx, py + dy]
        for (px, py), (dx, dy) in zip(base_corners, offsets)
    ]

    # Create draw_line actions
    actions = []
    for i in range(4):
        x0, y0 = perturbed[i]
        x1, y1 = perturbed[(i + 1) % 4]  # Wrap around to close the square
        actions.append(Action("draw_line", [x0, y0, x1, y1]))

    return actions

def gen_initial_guess(initial: DrawState):
    return {
        "pos": [0.32, 0.24], # Center of the drawing board (cm)
        "size": .2, # Square side length (cm)
        "point_offsets": [[0., 0.]] * 4 # 2D point offsets for the square (4 points)
    }
...
\end{lstlisting}

\subsection{Prompt for Push Environment}

\begin{lstlisting}

You are a franka panda robot operating in an environment with the following state:

TABLE_BOUNDS = [[-0.5, 0.5], [-0.5, 0.5], [0, 0]]  # X Y Z
TABLE_CENTER = [0, 0, 0]

@dataclass
class Frame:
    name: str
    x_pos: float
    y_pos: float
    z_pos: float
    x_rot: float
    y_rot: float
    z_rot: float
    size: float | list[float]
    color: list[float]

@dataclass
class PushState(State):
    frames: List[Frame] = field(default_factory=list)

    def getFrame(self, name: str) -> Frame:
        for f in self.frames:
            if f.name == name:
                return f
        return None

You have access to the following set of skills expressed as pddl predicates followed by descriptions. 
You have no other skills you can use, and you must exactly follow the number of inputs described below.
The coordinate axes are x, y, z where x is left/right from the robot base, y the distance from the robot base, and z is the height off the table.

Action("pick", [frame_name], pick_axis)
Grab the frame with name frame_name. Aligns the gripper x-axis (the axis in which the fingers move) with the pick_axis, if set to None, the all axis are checked and which ever one is feasible gets used.

Action("place_sr", [x, y, z, rotated, yaw])
Place grasped object at pose x, y, z, with a specific yaw angle. If the rotated boolean is set to True, it will rotate the block 90 degrees around the pick axis. The yaw, which is in radians, determines the angle at which the object is rotated with respect to the object's current local axis pointing upwards. If z is set to None, the object gets plazed on the table. If yaw is set to None, there are no restrictions to the yaw angle.

Action("push_motion", [start_x, start_y, end_x, end_y])
Perform a push motion along the straight 2D path defined by the start and end points.

Your goal is to generate two things:

First, generate a python function named `gen_plan` that can take any continuous inputs. No list inputs are allowed.
and return the entire plan with all steps included where the parameters to the plan depend on the inputs.

Second, generate a python function `gen_initial_guess` that returns a set of initial guesses for the continuous input parameters. The number of initial guesses in the
`gen_initial_guess` should exactly match the number of inputs to the function excluding the state input.

The main function should be named EXACTLY `gen_plan` and the initial_guess of the main function should be named EXACTLY `gen_initial_guess`. Do not change the names. Do not create any additional classes or overwrite any existing ones.
Aside from the inital state all inputs to the `gen_plan` function MUST NOT be of type List or Dict. List and Dict inputs to `gen_plan` are not allowed.

Additionally, the input to `gen_initial_guess` must be exactly the `initial:PushState` argument, even if this isn't explicitly used within the function!

#define user
initial=PushState(frames=[Frame(name="block_red", x_pos=0.0, y_pos=0.0, z_pos=0.71, x_rot=-0.0, y_rot=0.0, z_rot=-0.0, size=[0.04, 0.04, 0.12, 0.0], color="[255, 0, 0]"), Frame(name="block_green", x_pos=0.15, y_pos=0.0, z_pos=0.71, x_rot=-0.0, y_rot=0.0, z_rot=-0.0, size=[0.04, 0.04, 0.12, 0.0], color="[0, 255, 0]"), Frame(name="block_blue", x_pos=0.3, y_pos=0.0, z_pos=0.71, x_rot=-0.0, y_rot=0.0, z_rot=-0.0, size=[0.04, 0.04, 0.12, 0.0], color="[0, 0, 255]"), Frame(name="l_gripper", x_pos=0.0, y_pos=0.28, z_pos=1.27, x_rot=0.5, y_rot=-0.0, z_rot=1.29, size=[0.03], color="[229, 229, 229]"), Frame(name="table", x_pos=0.0, y_pos=0.0, z_pos=0.6, x_rot=-0.0, y_rot=0.0, z_rot=-0.0, size=[1.0, 1.0, 0.1, 0.02], color="[76, 76, 76]")])
Goal: Build a bridge. A bridge is defined as two vertical blocks next to each other and one horizontal block on top of them.

#define assistant
```python
def gen_plan(state: PushState, center_x: float, center_y: float, yaw: float, slack: float):
    
    import numpy as np
    
    # Build the bridge
    actions = []

    block_size_z = state.getFrame("block_red").size[2]
    
    # Red #
    pos_x = np.cos(yaw) * block_size_z * .5 + center_x
    pos_y = -np.sin(yaw) * block_size_z * .5 + center_y
    actions.append(Action("pick", ["block_red", None]))
    actions.append(Action("place_sr", [pos_x, pos_y, None, None, None]))

    # Green #
    pos_x = -np.cos(yaw) * block_size_z * .5 + center_x
    pos_y = np.sin(yaw) * block_size_z * .5 + center_y
    actions.append(Action("pick", ["block_green", None]))
    actions.append(Action("place_sr", [pos_x, pos_y, None, None, None]))

    # Blue #
    pos_z = block_size_z + slack
    actions.append(Action("pick", ["block_blue", None]))
    actions.append(Action("place_sr", [center_x, center_y, pos_z, True, yaw]))
    
    return actions

def gen_initial_guess(initial:PushState):
    guess = {
        "center_x": .2, # BBO initial value
        "center_y": .2,
        "yaw": .0,
        "slack": .03,
    }
    return guess
```

#define user
initial=PushState(frames=[Frame(name="big_red_block", x_pos=-0.2, y_pos=0.3, z_pos=0.7, x_rot=-0.0, y_rot=-0.0, z_rot=1.57, size=[0.1, 0.2, 0.1, 0.0], color="[204, 51, 63]"), Frame(name="target_pose", x_pos=0.4, y_pos=0.3, z_pos=0.7, x_rot=-0.0, y_rot=0.0, z_rot=-2.51, size=[0.1, 0.2, 0.1, 0.0], color="[0, 255, 0]")])
Goal: Push the red block to the taget pose.

#define assistant
```python
def gen_plan(state: PushState,
             a_start_x_offset: float, a_start_y_offset: float,
             a_end_x_offset:   float, a_end_y_offset:   float,
             b_start_x_offset: float, b_start_y_offset: float,
             b_end_x_offset:   float, b_end_y_offset:   float):
    
    import numpy as np

    # Build the block towards the target
    actions = []
    red_box = state.getFrame("big_red_block")
    target = state.getFrame("target_pose")
    dir = np.array([target.x_pos, target.y_pos]) - np.array([red_box.x_pos, red_box.y_pos])
    dir_normed = dir / np.linalg.norm(dir)

    # First push start
    offset_mag = max(red_box.size[:2]) * 3
    a_start_x = red_box.x_pos - dir_normed[0]*offset_mag + a_start_x_offset
    a_start_y = red_box.y_pos - dir_normed[1]*offset_mag + a_start_y_offset
    
    # First push end
    a_end_x = target.x_pos + a_end_x_offset
    a_end_y = target.x_pos + a_end_y_offset
    
    # Second push start
    b_start_x = target.x_pos - dir_normed[0]*.2 + a_end_x_offset
    b_start_y = target.y_pos - dir_normed[1]*.2 + a_end_y_offset
    
    # Second push end
    b_end_x = target.x_pos + a_end_x_offset
    b_end_y = target.x_pos + a_end_y_offset

    # First Push #
    actions.append(Action("push_motion", [a_start_x, a_start_y, a_end_x, a_end_y]))

    # Second Push (For adjusting position) #
    actions.append(Action("push_motion", [b_start_x, b_start_y, b_end_x, b_end_y]))

    return actions

def gen_initial_guess(initial: PushState):
    return {
        "a_start_x_offset": .0, # BBO initial value
        "a_start_y_offset": .0,
        "a_end_x_offset": .0,
        "a_end_y_offset": .0,
        "b_start_x_offset": .0,
        "b_start_y_offset": .0,
        "b_end_x_offset": .0,
        "b_end_y_offset": .0,
    }
\end{lstlisting}